\RequirePackage{fix-cm}

\RequirePackage{snapshot}

\documentclass[twocolumn]{svjour3}          
\smartqed  

\newcommand{\minisection}[1]{\vspace{0.04in} \noindent {\bf #1}\ \ }
\usepackage{subcaption}
\usepackage{multirow}
\usepackage{multicol}

\usepackage{amsmath,amsfonts,bm}

\def\eqref#1{equation~\ref{#1}}

\def\1{\bm{1}}

\DeclareMathAlphabet{\mathsfit}{\encodingdefault}{\sfdefault}{m}{sl}
\SetMathAlphabet{\mathsfit}{bold}{\encodingdefault}{\sfdefault}{bx}{n}

\usepackage{graphicx}%
\usepackage{amsmath,amssymb,amsfonts}%
\usepackage{amsthm}%
\usepackage{mathrsfs}%
\usepackage[title]{appendix}%
\usepackage{xcolor}%
\usepackage{textcomp}%
\usepackage{manyfoot}%
\usepackage{booktabs}%
\usepackage{algorithm}%
\usepackage{algorithmicx}%
\usepackage{algpseudocode}%
\usepackage{listings}%
\usepackage{url}
\usepackage{capt-of}
\usepackage{multirow}
\usepackage{makecell}
\usepackage{wrapfig}
\usepackage{array}
\usepackage{amssymb}
\usepackage{pifont}%
\newcommand{\cmark}{\ding{51}}%
\newcommand{\xmark}{\ding{55}}%

\usepackage{soul, xcolor}
\setstcolor{red}

\usepackage{array}
\makeatletter
\newcommand*{\@rowstyle}{}
\newcommand*{\rowstyle}[1]{
  \gdef\@rowstyle{#1}%
  \@rowstyle\ignorespaces%
}

\newcolumntype{=}{
  >{\gdef\@rowstyle{}}%
}

\newcolumntype{+}{
  >{\@rowstyle}%
}

\makeatother

\usepackage[round]{natbib}
\usepackage[final, colorlinks=true,allcolors=blue,breaklinks=true]{hyperref} 

\definecolor{f_trail}{RGB}{170,170,170}
\definecolor{f_grass}{RGB}{0,255,0}
\definecolor{f_vegetation}{RGB}{102,102,51}
\definecolor{f_sky}{RGB}{0,120,255}
\definecolor{f_obstacle}{RGB}{0,0,0}
\definecolor{u_Bed}{RGB}{0,0,255}
\definecolor{u_Books}{RGB}{233, 89, 48}
\definecolor{u_Ceiling}{RGB}{0, 218, 0}
\definecolor{u_Chair}{RGB}{149, 0, 240}
\definecolor{u_Floor}{RGB}{222, 241, 24}
\definecolor{u_Furniture}{RGB}{255, 206, 206}
\definecolor{u_object}{RGB}{0, 224, 229}
\definecolor{u_Picture}{RGB}{106, 136, 204}
\definecolor{u_Sofa}{RGB}{117, 29, 41}
\definecolor{u_Table}{RGB}{240, 35, 235}
\definecolor{u_Tv}{RGB}{0, 167, 156}
\definecolor{u_wall}{RGB}{250, 139, 0}
\definecolor{u_Window}{RGB}{225, 229, 195}

\begin{document}
\sloppy
\title{Exemplar-free Continual Learning of Vision Transformers via Gated Class-Attention and Cascaded Feature Drift Compensation}

\author{Marco Cotogni,
        Fei Yang,
        Claudio Cusano,\\
        Andrew D. Bagdanov,
        Joost van de Weijer 
}
\titlerunning{Exemplar-free Continual Learning of ViTs via Gated Class-Attention and Cascaded Feature Drift Compensation}
\authorrunning{Cotogni et al.} 

\institute{M. Cotogni and C. Cusano are with the Dept. of Electrical, Computer and Biomedical Engineering, University of Pavia, Pavia, Italy. \email{marco.cotogni01@universitadipavia.it, claudio.cusano@unipv.it}. \\
F. Yang and J. van de Weijer are with the Computer Vision Center, Universitat Aut\`onoma de Barcelona, Barcelona 08193, Spain. \email{\{fyang, joost\}@cvc.uab.es}. \\
A. D. Bagdanov is with MICC, University of Florence Florence 50134, FI, Italy. \email{andrew.bagdanov@unifi.it}\\
}


\maketitle

\begin{abstract}
 Vision transformers (ViTs) have achieved remarkable successes across a broad range of computer vision applications. As a consequence there has been increasing interest in extending continual learning theory and techniques to
    ViT architectures. We propose a new method for exemplar-free class incremental training of ViTs. The main challenge of exemplar-free continual learning is maintaining plasticity of the learner without causing catastrophic forgetting of previously learned tasks. This is often achieved via exemplar replay which can help recalibrate previous task classifiers to the feature drift which occurs when learning new tasks. Exemplar replay, however, comes at the cost of retaining samples from previous tasks which for many applications may not be possible.  
    
    To address the problem of continual ViT training, we first propose \emph{gated class-attention} to minimize the drift in the final ViT transformer block. This mask-based gating is applied to class-attention mechanism of the last transformer block and strongly regulates the weights crucial for previous tasks. Importantly, gated class-attention does not require the task-ID during inference, which distinguishes it from other parameter isolation methods. Secondly, we propose a new method of \emph{feature drift compensation} that accommodates feature drift in the backbone when learning new tasks. The combination of gated class-attention and cascaded feature drift compensation allows for plasticity towards new tasks while limiting forgetting of previous ones. Extensive experiments performed on CIFAR-100, Tiny-ImageNet and ImageNet100 demonstrate that our exemplar-free method 
    obtains competitive results when compared to rehearsal based ViT methods.\footnote{Code: \url{https://github.com/OcraM17/GCAB-CFDC}}
\end{abstract}


\section{Introduction}\label{sec1}
The initial excellent results of transformers for language tasks~\citep{vaswani2017attention} have encouraged its application also for vision applications~\citep{dosovitskiy2020image}. Vision Transformers (ViTs) currently achieve excellent results for many applications~\citep{caron2021emerging,liu2021swin,strudel2021segmenter}. Most existing work on ViT training assumes that all training data is jointly available, an assumption which does not hold for real-world applications in which data arrives in a sequence of non-overlapping tasks. Continual learning considers learning from a non-IID stream of data. Applying a naive finetuning approach to such data results in a phenomenon called \emph{catastrophic forgetting} which results in a drastic drop in performance on previous tasks~\citep{goodfellow2014empirical}. The main goal of continual learning algorithms is to maximize the stability-plasticity trade-off \citep{mermillod2013stability}, i.e. to mitigate forgetting of previously learned classes while maintaining the plasticity required to learn new ones. 

\begin{figure}
    \centering
    \includegraphics[width=\linewidth]{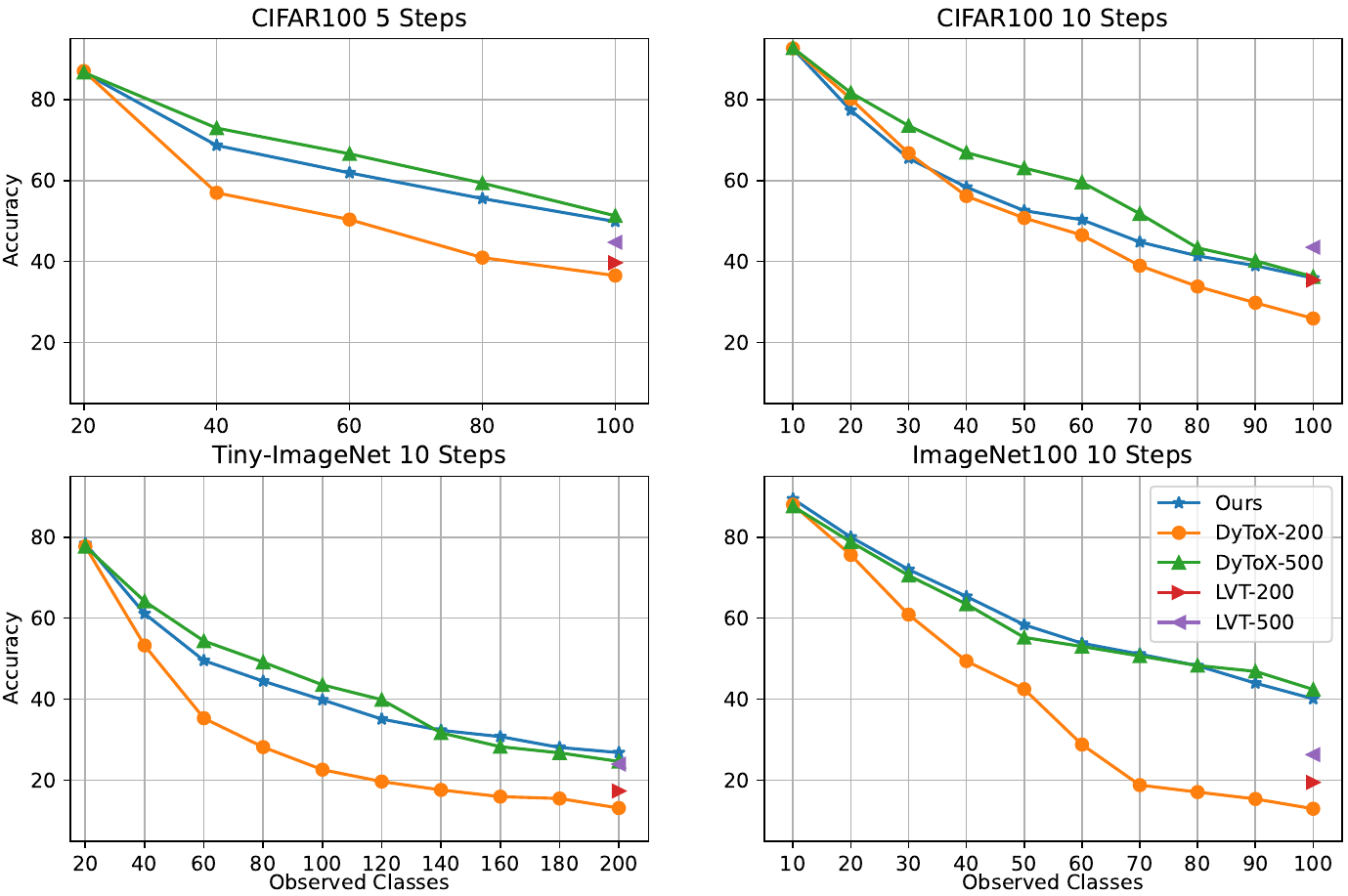}
    \vspace{-0.4cm}
    \caption{Comparison between GCAB and transformer-based state-of-the-art methods~\citep{douillard2022dytox, wang2022continual} (number refers to number of exemplars in replay buffer). Our exemplar-free method obtains comparable performance compared to rehearsal-based methods on all scenarios. LVT results are only reported after the last task, since no code has been released.}
    \label{fig:allscenarios}
\end{figure}

One of the most successful approaches to preventing forgetting of previous tasks is \emph{exemplar rehearsal} in which a subset of images from previous tasks is stored and then rehearsed when learning new ones~\citep{bang2021rainbow,buzzega2020dark,buzzega2021rethinking, chaudhry2018efficient, riemer2018learning, rebuffi2017icarl, lopez2017gradient}. Because of its success, the rehearsal technique has also been adopted by the initial works on continual learning for ViTs~\citep{douillard2022dytox,wang2022continual}. However, for many applications the storage of previous task data might not be possible. This is especially true for applications with strict memory constraints and those where privacy or data use legislation prevents the long-term storage of data. In order to overcome this limitation, exemplar-free methods have been investigated~\citep{kirkpatrick2017overcoming,li2017learning,yan2021dynamically,yu2020semantic}. These methods do not store any data from previous tasks, however their application to continual learning of ViTs has not been fully explored. 

In this paper, we propose one of the first exemplar-free ViT-based methods for class-incremental learning (CIL).  
One of the challenges of exemplar-free continual learning is that these models tend to forget previously learned features while they are learning new tasks. 
The architecture is based on a \textbf{gated class-attention} mechanism applied to the ViT decoder in order to mitigate the forgetting of learned features. Existing mask-based gating methods have the major drawback that they can only be applied to task-incremental learning, since they require the task-ID at inference time~\citep{abati2020conditional, del2020ratt, serra2018overcoming}. 
We propose a solution to overcome this limitation by applying the masking mechanism only on the transformer decoder via multiple forward passes.
This solution allows us to use mask-based gating for class-incremental learning without the need for task-IDs at inference time.

Mask-based gating prevents the drift of weights in the decoder, however it does not mitigate the drift in the transformer encoder. In fact, learning a stable backbone in the exemplar-free scenario is very difficult due to the drift in encoder weights that occurs during the learning of new tasks. To address this, we propose a method for \emph{backbone regularization in combination with a feature drift compensation} mechanism that uses a cascade of projection networks that map the current backbone features to those of the previous backbone. This allows increased plasticity while maintaining stability across tasks, incurring a small computational cost due to the feature projection cascade. We also show, however, that knowledge distillation can be used to alleviate the computational burden of the projection cascade and the need for multiple forward passes at inference time.

The main contributions of this work are:
\begin{itemize}
    \item a gated class-attention mechanism, called GCAB, that mitigates weight drift in the transformer decoder, which is the first application of an activation-masking method for class-incremental learning; 
    \item a novel method for feature drift compensation that increases plasticity towards new classes while maintaining the stability of previously learned ones;
    \item a method for GCAB distillation that reduces the computational overhead due to multiple forward passes and the memory overhead of storing projection networks;
    \item experiments on multiple benchmarks demonstrate that our exemplar-free approach achieves state-of-the-art performance compared to other exemplar-free methods and outperforms recent continual learning methods developed for ViT even when these are equipped with a small memory of exemplars (see Figure~\ref{fig:allscenarios}).
\end{itemize}

\section{Related Work}

\minisection{Continual Learning.} Continual learning algorithms can be grouped in three categories \citep{delange2021continual}: regularization approaches, parameter based regularization~\citep{kirkpatrick2017overcoming, lee2020continual,liu2018rotate,zenke2017continual}, and data based regularization~\citep{castro2018end, dhar2019learning, hou2019learning,jung2016less,li2017learning,  wu2019large,zhang2020class}; rehearsal approaches, which store~\citep{chaudhry2018riemannian,rebuffi2017icarl} or generate exemplars~\citep{wang2021ordisco,zhai2021hyper}; and bias-correction approaches~\citep{castro2018end,hou2019learning,wu2019large}. 

 \minisection{Continual Learning with VITs.} 
 Visual Transformers recently outperforms convolutional neural networks and in particular resnet~\citep{he2016deep} in several tasks like classification~\citep{dosovitskiy2020image} or segmentation~\citep{zheng2021rethinking}. 
Although ViTs are considered state-of-the-art models, their application in continual learning has not been fully explored. \cite{douillard2022dytox} proposed a transformer-based architecture, called DyTox, with a dynamic task-token expansion for mitigating catastrophic forgetting. \cite{wang2022continual} proposed an inter-task attention mechanism for ViTs. \cite{wang2022learning} described a prompting method for continually learning a classifier using a pretrained, frozen ViT backbone.
Even if the performances showed in these work are very remarkable, the challenge of continually learning the parameters of a ViT without storing exemplars or using a pretrained model, is still open. 
Differently from previous work, we propose an exemplar-free ViT approach to class-incremental learning. Our work is inspired by DyToX~\citep{douillard2022dytox}, which applies a task conditioned class-attention block. However, DyToX shares the class-attention block parameters between tasks which leads to forgetting, and it therefore requires exemplars to counter this. Instead, we replace the task token with a task-specific gating function that prevents forgetting and does not require exemplars. In addition, we introduce feature drift compensation, which allows for more plasticity in the backbone.

\minisection{Parameter Isolation in Continual Learning.} In this family of algorithms, learnable masks are applied to the weights of the model in order to reduce forgetting.
\cite{mallya2018piggyback} proposed Piggyback, a masked-based method able to learn the weight masks while training a backbone. The same group proposed Packnet~\citep{mallya2018packnet} which, via iterative pruning and sequential re-training, is able to add multiple tasks to a single network. \cite{serra2018overcoming} proposed to apply masks to layer activations in order to limit the update of the parameters more relevant to a specific task. \cite{masana2021ternary} proposed a system of ternary-masks applied on to layer activations for preventing catastrophic forgetting and backward transfer. \cite{yan2021dynamically} proposed a Dynamical Expandable Representation (DER) for continual learning. In this work, channel-level masks are used for pruning the feature extractor.
\cite{rajasegaran2019random} proposed Random Path Selection (RPS). This approach uses a parameter isolation mechanism, distillation, and a replay-buffer to learn different paths for each task without the need for a task-ID during inference.

\begin{figure}
    \centering
    \includegraphics[width=.95\linewidth]{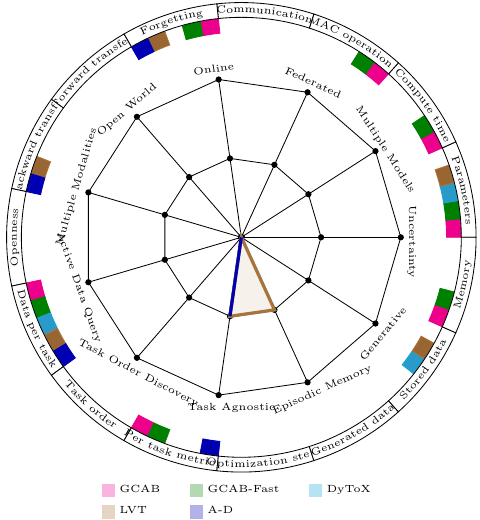}
    \caption{CLEVA-Compass for transformer-based methods (zoom in for a better view). In the inner circle, GCAB, GCAB-Fast are covered by A-D, while Dytox is covered by LVT. We do not report backward transfer since it makes more sense for task-incremental learning.}
    \label{fig:cleva}
\end{figure}

\begin{figure*}[tb]
\begin{center}
\includegraphics[width=0.9\textwidth]{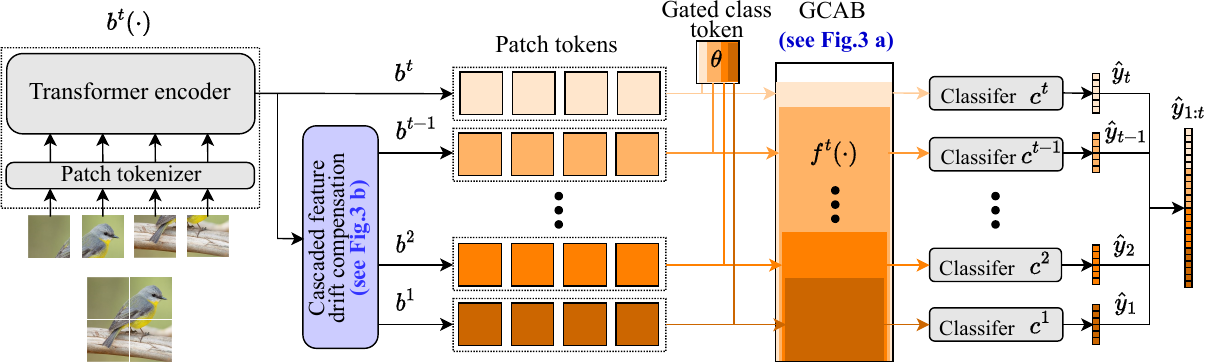}
\end{center}
\caption{Overview of the architecture. Our main contributions are the \emph{gated class-attention block (GCAB)} to prevent forgetting in the final ViT block (Section~\ref{sec:gated-class-attention}) and the \emph{cascaded feature drift compensation} to compensate for feature drift of the backbone network (Section~\ref{sec:distillation}).}
\label{fig:trasf}
\end{figure*}

\minisection{Exemplar-Free Continual Learning.} This is one of the most challenging scenario in continual learning. In this paradigm, it is not possible to store any exemplars from the previously observed classes. \cite{li2017learning} proposed an exemplar-free data regularization approach to mitigate forgetting. This method distills knowledge of the previous model into the new one in order to prevent weight drift while learning the new task data. \cite{kirkpatrick2017overcoming} described an exemplar-free  weight regularization approach called Elastic Weight Consolidation (EWC) for preventing weight drift. Similarly, \cite{aljundi2018memory} proposed a method that accumulates the importance of each model parameter by analyzing the effect of their change to the predicted output. \cite{yu2020semantic} proposed a semantic drift compensation mechanism to compensate for feature drift in previous tasks by approximating it with the drift estimated with current task data. \cite{toldo2022bring} presented a framework for modeling the semantic drift of model weights of and estimating feature drift in the representation of previously learned classes. \cite{pelosin2022towards} proposed an attention distillation mechanism for exemplar-free visual transformer in task-incremental learning.

In Figure \ref{fig:cleva} we report the CLEVA-Compass \cite{mundt2021cleva} comparing our method, GCAB and GCAB-Fast, to other transformer methods (A-D, DyTox, and LVT) on several measures.

\section{Method}
\label{sec:method}

\subsection{Problem setup}
Here we define the class-incremental learning setup and the specific Vision Transformer architecture we use.

\minisection{Class-incremental learning setup.} In class-incremental learning the model learns a sequence of $T$ tasks, where each task $t$ introduces a number of new classes $C^{t}$.  The data $D^t$ of task $t$ contains samples $\left(x_{i}, y_i\right)$, where $x_{i}$ is input data labeled by $y_{i} \in C^t$. Note that we consider the case in which that there is no overlap between different task label sets: $C^i\!\cap\!C^j\!=\!\emptyset$ if $i\!\neq\!j$, as is commonly assumed~\citep{masana2020class}. The model is evaluated on all previously seen classes $C^{\le t}=\cup_{t' \leq t} C^{t'}$. Class-incremental learning differs from task-incremental learning in that it has no access to the task label $t$ at inference time, and is therefore considered a more challenging setting~\citep{delange2021continual}. Furthermore, we consider the more restrictive setup of exemplar-free incremental learning in which no data from previous tasks is saved. 

\minisection{Transformer architecture.} 
We use a vision transformer based on the one proposed by \cite{dosovitskiy2020image} and the recent improvements of~\cite{touvron2021going}. It consists of a transformer encoder and decoder, each built with several multi-head attention blocks. In Figure~\ref{fig:trasf} we give a schematic diagram of our architecture. Formally, the input image $x \in \mathbb{R}^{H \times W \times C}$ is passed through a patch tokenizer that splits $x$ into $N$ patches and projects them using a 2D convolutional layer to obtain a set of $N$ patch tokens $x_0\in \mathbb{R}^{N \times D}$. A learnable position embedding 
is added to the patch tokens as in~\citep{gehring2017convolutional}. The patch tokens $x_0$ are passed as input to a sequence of $M$ transformer encoder blocks, each yielding tensors of the same dimensions. Each block is composed of a multi-head self-attention (SA) mechanism \citep{vaswani2017attention}, layer normalization and a Multi-layer Perceptron (MLP), each with residual connections: 
 \begin{equation}
    \begin{aligned}
        x'_l&=x_l+\text{SA}(x_l)\\
        x_{l+1}&= x'_l+\text{MLP}(x'_l)
        \label{eq:ViTblock}
     \end{aligned}
 \end{equation}
We follow the design of CaiT~\citep{touvron2021going}, and only insert a class token with a class-attention layer in the last block of the decoder. The transformer decoder is composed of one single block. To distinguish the various parts of the transformer network, we define the image output prediction $\hat y=c(f(b(x; \Psi)))$, where the backbone features $b(x; \Psi) \in \mathbb{R}^{N\times D}$ parameterized by $\Psi$ are the output of the self-attention blocks 
, and $f(b(x; \Psi))$ refers to the feature output of the decoder
before classifier $c$.

\begin{figure*}[tb]
\begin{center}
\includegraphics[width=\linewidth]{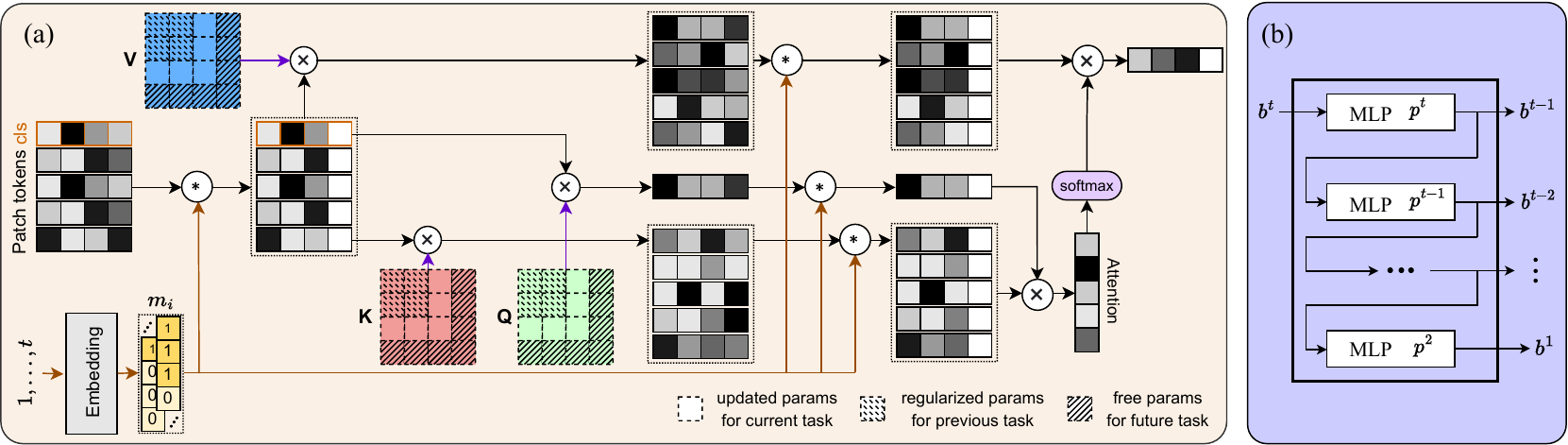}
\end{center}
\caption{(a) The gated self-attention layer (for a single head) in GCAB. (b) Feature Drift Compensation applies a cascade of feature projection networks to map the current backbone features to previous task features to compensate for feature drift.}
\label{fig:gating_proj}
\end{figure*}

\subsection{Gated class-attention}
\label{sec:gated-class-attention}

Parameter isolation methods work by isolating a limited set of parameters after learning each task~\citep{delange2021continual, mallya2018packnet, rusu2016progressive,serra2018overcoming}.
Inspired by~\cite{serra2018overcoming}, we design a parameter isolation method called \emph{gated class-attention} that operates on the class attention block of transformers.  
The method allows parameters of the network used by previous tasks to be exploited by new tasks, thereby allowing for forward transfer, but their update is restricted to prevent forgetting. The main strengths of this approach are the good forward transfer with little or no forgetting of previous tasks, together with the ability to automatically learn which neurons to dedicate to each task within the capacity limit of the neural network.

The forward pass of parameter isolation methods is usually conditioned on the task~\citep{mallya2018packnet,masana2021ternary,serra2018overcoming}, and is therefore restricted to task-incremental learning. A possible way to extend these methods to class-incremental learning would be to run one forward pass for each task and then combine the task predictions (e.g., by concatenation). However, this would increase the run time linearly in the number of tasks. To limit computational overhead, we propose to only apply attention masks to the last block of the ViT. In Section~\ref{sec:distillation}, we investigate distillation to further reduce computational overhead.

\minisection{Mask-based class-attention gating.} We apply the attention-gating in the last transformer block, i.e. the decoder, which contains class-attention~\citep{touvron2021going}. This block combines the patch tokens from previous blocks with a learnable class token $\theta$. 

The learnable masks $m$ for task $t$ are defined as:
\begin{eqnarray}
\label{eq:mask}
m^t=\sigma(sAt^{\textrm{T}}),
\end{eqnarray}
where, slightly abusing notation, $t$ represents both a task \emph{index} and a \emph{one-hot vector} identifying the current task, $A$ is a learnable embedding matrix, $\sigma$ is the sigmoid activation function, and $s$ is a positive scaling parameter. 
Here $m^t$ refers to the neurons that have been selected for task $t$.

We can now define the forward pass through the final class-attention block by introducing a mask for all the activations contributing to the final class token output. These masks correspond to: the input tokens ($m^t_i$), queries and keys ($m^t_{QK}$), values ($m^t_V$), the MLP ($m_1$ and $m_2$), and the class-attention output ($m_o$). We can then compute the query ($Q$), key ($K$), value ($V$), attention ($A$) and self attention output ($O$) given the block token inputs $O^t$ and these masks:
\begin{align}
    p &=[\theta, b] \in \mathbb{R}^{(N+1)\times D} \nonumber \\
    Q^t &=W_q(\theta\odot m^t_i) \nonumber \\
    K^t &=W_k( p \odot m^t_i) \nonumber \\
    V^t &=W_v( p \odot m^t_i) \\
    A^t &=\text{softmax}\left\{ \left(\left(Q^t \odot m^t_{QK}\right) \left(K^t\odot m^t_{QK}\right)^{T}\right) / \sqrt{d/h}\right\} \nonumber \\
    O^t &=W_o A^t(V^t\odot m^t_{V}) \nonumber.
\label{eq:SA}
\end{align}

The gating mechanism is also applied to the MLP (see Eq.~\ref{eq:ViTblock}) according to:
\begin{equation}
\begin{aligned}
     b'^t &= O^t  + \theta \\
     u^t &= W_1(b'^t \odot m^t_1) \\
     v^t &= W_2(u^t \odot m^t_2) \\
     f^t &= v^t + O^t
\end{aligned}
\label{eq:MLP}
\end{equation}
We call this the Gated Class-Attention Block (GCAB). The masks are learned during the training of task $t$ and their role is twofold: they select those activations that are used to compute the task-conditioned output and they restrict the backpropagation of future tasks, preventing changes to weights that used by previous tasks. 
In the experimental section, we show that good results are obtained when sharing the weights and setting $m_{QK}=m_V=m_1=m_o=m_i$, resulting in only two learnable masks $m_i$ and $m_2$. In Figure~\ref{fig:gating_proj} (a) we show the interaction between the masks $m_i$ and self-attention layer in the class-attention block.

For each task a dedicated classifier $c^t$ is added  which produces predictions $\hat y^t = W_{clf} (f^t\odot m^t_1)$ using the vector $f^t$ output from the GCAB.  We perform multiple forward passes of patches extracted from the backbone $b^t$ through the decoder $f^t$ (i.e. we pass it $t$ times). For each pass $s$, the obtained vectors which we denote by $f^t_s$ (and are computed with the masks $m^s$), are then passed to the corresponding classifier $c^s$. The classifier outputs are then concatenated $C=[c^1,\cdots,c^t]$ and the binary cross entropy loss $\mathcal{L}_{\text{BCE}}$ is computed.

\begin{figure*}
\begin{center}
\includegraphics[width=\linewidth]{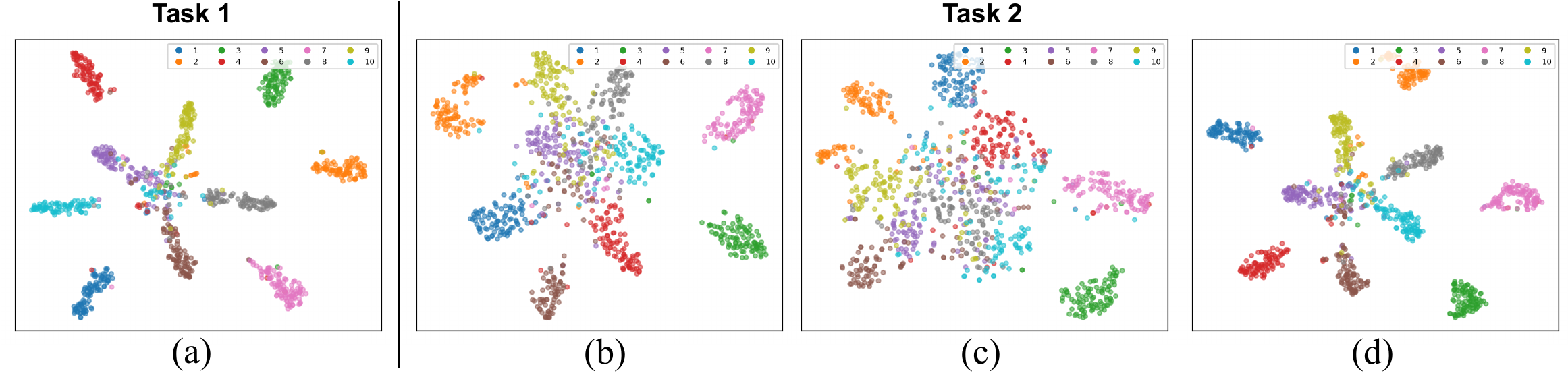}
\end{center}
\caption{t-SNE visualization of the output of the decoder $f(\cdot)$ after the first task and second task.
These visualizations are based on the task-agnostic embeddings in the 10 task scenario on CIFAR-100.
(a) Results after task 1. Results after task 2 (b) without PFR during training, (c) with PFR during training but without feature drift compensation, and (d) with PFR during training and with feature drift compensation.
More t-SNE results are provided in the Appendix B.
}
\label{fig:embedding_tag_red}
\end{figure*}

\minisection{Training.}
During the training of the current task $t$ 
the scaling parameter $s$ from equation~\ref{eq:mask}, is scaled with the batch index: $s = \frac{1}{s_{max}} + (s_{max}-\frac{1}{s_{max}})\frac{i-1}{I-1}$, where $i$ current batch index and $I$ is the total number of batches in an epoch. This was found to be beneficial in~\cite{serra2018overcoming}.

At the end of the current task, the learned masks are accumulated as $m_*^{<t} = \max (m_*^{t-1},m_*^{<t-1})$, where '$*$' stands for any of the specific mask subscripts introduced above. Accumulated masks $m_*^{<t}$ are used during backpropagation to prevent updating weights considered important for the tasks observed so far. The masks are learned by minimizing the following loss function:
\begin{equation}
     \mathcal{L}_{\text{GCAB}} = \lambda_{\text{GCAB}} \frac{\sum_{x} m^{t}_{x}(1-m^{<t}_{x})}{\sum_{x}(1-m^{< t}_{x})},
\end{equation}
where $m_{x}^{t}$ is the mask learned at the current task $t$ for component $x$ of the GCAB, $x$ ranges over the mask subscripts described above, and $m^{<t}_{x}$ is the cumulative mask. $\lambda_{\text{GCAB}}$ is a tunable hyperparameter controlling the capacity of the masks learned during the tasks. This equation encourages new task mask $m_x^t$ to be sparse, however it permits use of activations already used by previous tasks $m_x^{<t}$ at no cost.

The cumulative masks play a pivotal role during the training of new tasks. Consider, for example, weights $W_q$ that map from the input tokens (masked by $m_i^t$) to the queries (masked by $m_{QK}^t$). We then define the elements of the weight mask according to $M^{<t}_{q,kl}=1-\min(m_{i,k}^{<t},m_{QK,l}^{<t})$ where $m_{i,k}^{<t}$ refers to the $k$-th element of $m_{i}^{<t}$. The update rule for the backpropagation of the gradient is then: $ W_{q} = W_{q} - \lambda M_{q}^{< t} \odot \frac{\partial L}{\partial W_{q}}$. This update rule prevents the updating of part of the weights learned for previous tasks. The input mask also influences the updating of the class token embedding which is given by $W_\theta =W_\theta-\lambda (1-m_{i}^{<t}) \odot \frac{\partial L}{\partial W_{\theta}}$. 

For completeness, we report here the update rules for all weight matrices in the GCAB:
\begin{center}
\resizebox{1.025\linewidth}{!}{\begin{tabular}{ll}%
    $W_{Q} = W_{Q} - \lambda M_{q}^{< t} \odot \frac{\partial L}{\partial W_{q}}$
  &
    $W_{k} = W_{k} - \lambda M_{k}^{< t} \odot \frac{\partial L}{\partial W_{k}}$ \\
    $W_{v} = W_{v} - \lambda M_{v}^{< t} \odot \frac{\partial L}{\partial W_{v}}$
  &
    $W_\theta =W_\theta-\lambda (1-m_{i,k}^{<t}) \odot \frac{\partial L}{\partial W_{\theta}}$ \\
    $W_{o} = W_{o} - \lambda M_{o}^{< t} \odot \frac{\partial L}{\partial W_{o}}$
  &
    $W_{1} = W_{1} - \lambda M_{1}^{< t} \odot \frac{\partial L}{\partial W_{1}}$ \\

    $W_{2} = W_{2} - \lambda M_{2}^{< t} \odot \frac{\partial L}{\partial W_{2}}$
 &
    $W_{\text{clf}} = W_{\text{clf}} - \lambda (1-m_{o}^{< t}) \odot \frac{\partial \text{clf}}{\partial W_{\text{clf}}}$.
\end{tabular}}
\end{center}
Moreover, the weight masks are defined as:
\begin{center}
\resizebox{1.025\linewidth}{!}{\begin{tabular}{ll}
    $M^{<t}_{q,kl}=1-\min(m_{i,k}^{<t},m_{QK,l}^{<t})$
  &
    $M^{<t}_{k,kl}=1-\min(m_{i,k}^{<t},m_{QK,l}^{<t})$ \\
    $M^{<t}_{v,kl}=1-\min(m_{i,k}^{<t},m_{V,l}^{<t})$
  &
    $M^{<t}_{o,kl}=1-\min(m_{V,k}^{<t},m_{1,l}^{<t})$ \\
    $M^{<t}_{1,kl}=1-\min(m_{1,i}^{<t},m_{2,l}^{<t})$
  &
    $M^{<t}_{2,kl}=1-\min(m_{2,i}^{<t},m_{o,l}^{<t})$.
\end{tabular}}
\end{center}

\subsection{Backbone Regularization and Cascaded Feature Drift Compensation}
\label{sec:backbone}
In the previous section, we applied gating only to the last transformer block to limit computational overhead. Gating ensures that only minimal changes occur to the weights relevant to previous tasks, however, the network can still suffer from forgetting due to backbone feature drift.  

A straightforward way to prevent forgetting in the backbone network via regularization is feature distillation~\citep{hou2019learning}. Feature distillation encourages backbone features at task $t$ to remain close to those at task $t\textit{-}1$, however, this was found to limit plasticity~\citep{douillard2020podnet}. To ensure stability without sacrificing plasticity, some recent works in continual learning of self-supervised representations have proposed to learn a projector between feature extractors~\citep{fini2022self, gomez2022continually}. The approach, called \emph{Projected Functional Regularization} (PFR), 
introduces a projection network $p^t$ that maps the current backbone features to those of the previous backbone. This allows the new backbone to learn new features without imposing a high regularization penalty as long as the new features can still be projected back to those of the previous backbone. The  loss function is given by:
\begin{equation}
\mathcal{L}_{\text{pfr}}=
\lambda_{\text{pfr}} \mathbb{E}_{x \sim \mathcal{D}^t }
\left[\mathcal{S} \left( p^t\left(b\left(x;\Psi^t\right)\right), b\left(x;\Psi^{t-1}\right) \right)\right]
\label{eq:loss_prj}
\end{equation}
where $\mathcal{S}$ is the cosine distance, $\lambda_{\text{pfr}}$ is a trade-off parameter, and $\Psi^t$ refers to the parameters of the backbone after learning task $t$.

The gained plasticity induced by PFR leads to a misalignment of the current backbone with previous class-attention layers and classifiers. This is not a problem during self-supervised learning~\citep{fini2022self, gomez2022continually}, however for supervised learning when a previously learned classifier is applied to the current backbone it becomes problematic. Therefore, as the second main contribution, we propose \emph{cascaded feature drift compensation} that extends the use of projected feature regularization to supervised settings. This is especially relevant for exemplar-free methods, where alignment with previous tasks is challenging due to the absence of replay data.

The regularization of Eq.~\ref{eq:loss_prj} results in backbone drift, and therefore $b^t \neq b^{t-1}$. However, since we have the projection matrix $p^t$ we can approximate the previous backbone according to $b^{t-1}\approx p^t(b^t)$. Continuing this, we perform \emph{cascaded feature drift compensation} (FDC) (see Figure~\ref{fig:gating_proj}(b)) which applies the projection networks from previous tasks in a \emph{cascade} to align the current backbone with the learned class-attention block at any previous task $s \le t$:
\begin{equation}
    \hat y_{s}=c^t(f^t_s(p^{s+1}(p^{\cdots}(p^{t-1}(p^t(b^t(x)))))).
\label{eq:projection}
\end{equation}
Note that here $f_s^t$ refers to the class attention  block output at task $t$ computed based on the masks $m^s$. The projection networks $p^t$ must be saved in this formulation in order to compute the projection cascade that compensates for backbone feature drift. In Section~\ref{sec:distillation} we show how knowledge distillation can be used to eliminate the need for projection networks $p^t$ at inference time. 

To illustrate the effectiveness of cascaded feature drift compensation, we analyzed the embeddings produced by the GCAB. In Figure~\ref{fig:embedding_tag_red} we visualize the embedding produced by the GCAB using t-SNE~\citep{van2008visualizing}. 
The embedding of the images from the test set of the first task are shown after the training of the first task and the second task.
Only applying PFR during training results in a latent space where classes no longer align with their location after task 1 (see Figure~\ref{fig:embedding_tag_red}(c)). This is problematic since the previously trained classifier of task 1 no longer aligns with them and will therefore suffer a significant drop in performance (as verified in our ablation study). After applying the cascaded feature drift compensation, the class distributions are mapped back to their original locations and align again with the classifier head (see Figure~\ref{fig:embedding_tag_red}(d)). In conclusion, this illustration shows that cascaded feature drift compensation can be a strong tool to recover from feature drift in the backbone. This is important since it allows for high plasticity during the continual learning process.

\subsection{Training Objective and Inference}

The final objective we use for incremental training of the model is the sum of the three loss functions:
\begin{equation}
    \mathcal{L} = \mathcal{L}_{\text{BCE}} + \mathcal{L}_{\text{pfr}} +  \mathcal{L}_{\text{GCAB}} 
\end{equation}
After training the current task, an evaluation phase is performed over all classes seen so far.  At task $t$, the model is tested for the tasks $t' \leq t$. All images are passed through the backbone to obtain tokens $b$. The tokens are passed $t$ times through the Gated Class-Attention Block. Based on the task index of the forward pass $t'$, the composition of the previously stored projection networks is used as in Equation~\ref{eq:projection} to align the current backbone features to those of previous task $t'$. In Figure~\ref{fig:gating_proj} (b) we give a schematic diagram of the Feature Drift Compensation mechanism during inference. During inference the parameter $s$ of Equation~\ref{eq:mask} is equal to $s_{max}$.

\subsection{GCAB Distillation}
\label{sec:distillation}
\begin{figure}
    \centering
    \includegraphics[width=0.42\textwidth]{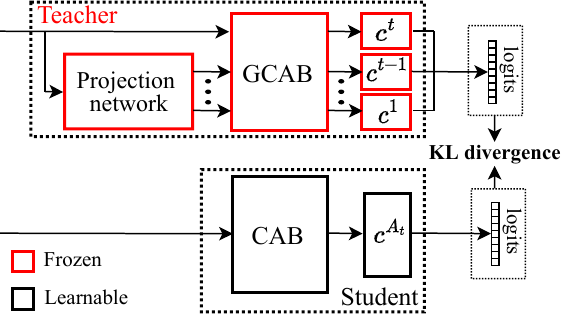}
    \caption{Illustration of GCAB distillation.}
    \vspace{-0.2cm}
    \label{fig:distill}
\end{figure}

To overcome the increased computational cost due to multiple forward passes for all tasks and the cascaded projection layers, we perform knowledge distillation~\citep{hinton2015distilling} to transfer the class-conditioned GCAB (the \emph{teacher})
into a single class-attention block (CAB) with the same architecture as the GCAB but \emph{without} masks, and an aggregated classifier $c^{A_{t}}$ (\emph{student}). 
{This reduces the number of parameters in the network, since the task projection networks are no longer needed, and eliminates the need for multiple forward passes. We name this distilled version  \emph{GCAB-Fast}}.
As shown in Figure~\ref{fig:distill}, the CAB and $c^{A_{t}}$ are trained by minimizing the Kullback–Leibler (KL) divergence  between the logits output by the teacher and student models. Note that this distillation is conducted only with the data from task $t$, while the transformer backbone $b$ and teacher hyper-classifier are frozen during training. We also investigate the use of static masks in the CAB to leave unused capacity in the student network in order to accommodate potential future tasks (see Section~\ref{sec:distillation-results} for details). 

\section{Experimental Results}

In this section we describe experimental results obtained with GCAB and GCAB-Fast. We begin with a description of our experimental setup and datasets used for comparison with the state-of-the-art.

\subsection{Experimental Setup and Datasets}
\label{sec:datasets}

Our network is based on the DytoX~\citep{douillard2022dytox} transformer architecture. The number of transformer encoder blocks $M=5$, each one with $H=12$ heads for the multi-head self-attention mechanism. The dimension of the embeddings is set to $D=384$. We use a 2-layer MLP for the projection network with also a dimensionality of 384 in the middle. We train each task for 500 epochs using Adam with $lr=1e^{-4}$ and a batch size of 128. We set hyperparameters $\lambda_{\text{pfr}}=0.001$, $\lambda_{\text{GCAB}}=0.05$ and $s_{max}=800$. 

For our experiments, we consider three datasets: CIFAR-100~\citep{krizhevsky2009learning}, Tiny-ImageNet~\citep{le2015tiny} and ImageNet100~\citep{russakovsky2015imagenet}. The CIFAR-100 dataset is composed of 60,000 images, each $32 \times 32$ pixels and divided into 100 classes. Each class has 500 training and 100 test images. The Tiny-ImageNet is a reduced version of the original ImageNet dataset with 200 classes. The classes are split into 500 for training, 50 for validation, and 50 for test (for a total number of 120,000 images). The images are $64 \times 64$ pixels.
The ImageNet100 dataset is a selection of 100 classes from the larger ImageNet dataset (composed of 1000 classes). The images in this reduced version are 120,000 for training and 5000 for test. For our purposes, we resized the images of this third dataset to $224 \times 224$.

\begin{table*}[tb]
\centering
  \begin{tabular}{lccccc}
    \hline
    \multirow{2}{*}{\textbf{Method}}
    &\multirow{2}{*}{\textbf{Exemplar-Free}}
    &\multicolumn{2}{c}{\textbf{CIFAR-100 5 Tasks}}
    &\multicolumn{2}{c}{\textbf{CIFAR-100 10 Tasks}}\\
    & &Class-IL &Task-IL &Class-IL &Task-IL\\
    \hline
    Joint& \cmark& $75.39$ & - & $75.39$ & -\\
    \hline
    ER \hfill \citep{riemer2018learning}   & \xmark &$21.94$ &$62.41$ &$14.23$ &$67.57$\\
    GEM \hfill \citep{lopez2017gradient} & \xmark&$19.73$& $57.13$&$13.20$ & $62.96$\\
    AGEM \hfill \citep{chaudhry2018efficient} & \xmark &$17.97$ &$53.55$ & \phantom{0}$9.44$  &$55.04$ 
     \\
    iCaRL \hfill \citep{rebuffi2017icarl} & \xmark&$30.12$ &$55.70$ &$22.38$ &$60.81$ 
     \\
    FDR \hfill \citep{benjamin2018measuring}  & \xmark &$22.84$ &$63.57$ &$14.85$ &$65.88$ 
     \\
    GSS \hfill \citep{aljundi2019gradient}  & \xmark &$19.44$ &$56.11$ &$11.84$ &$56.24$ 
     \\
    DER++ \hfill \citep{buzzega2020dark} & \xmark &$27.46$ &$62.55$ &$21.76$ &$59.54$ 
     \\
    HAL \hfill \citep{chaudhry2021using}  & \xmark &$13.21$ &$35.61$ & \phantom{0}$9.67$ &$37.49$ 
    \\
    ERT \hfill \citep{buzzega2021rethinking}  & \xmark &$21.61$ &$54.75$ &$12.91$ &$58.49$ 
     \\
    RM \hfill \citep{bang2021rainbow}   &\xmark &$32.23$ &$62.05$ &$22.71$ &$66.28$ 
     \\
    PASS \hfill \citep{zhu2021prototype} & \cmark & $48.28$ & - & $33.76$ & -\\
    SDC \hfill \citep{yu2020semantic}
    & \cmark& \phantom{0}6.65 &-& \phantom{0}7.41&-\\
    \hline
    LVT$^{\dag}$ \hfill \citep{wang2022continual}&\xmark &$39.68$ &$66.92$ &$35.41$ &$72.80$  
     \\
    DyToX$^{\dag}$ \hfill \citep{douillard2022dytox} &\xmark & $36.52$ & -  & $25.94 $ & -  
     \\
    A-D$^{\dag}$ \hfill \citep{pelosin2022towards}
    & \cmark & 15.61 & 42.82 & $16.77$& $55.53$\\
    \hline
    \textbf{GCAB}$^{\dag}$ & \cmark& $\mathbf{49.86 }$ & $\mathbf{81.01 }$ & $\mathbf{35.90 }$ & $\mathbf{82.08 }$ 
    \\
    \textbf{GCAB-Fast}$^{\dag}$ & \cmark & ${48.85 }$ & ${79.75 }$ & ${35.42 }$ & ${81.97 }$ 
    \\
    \hline
  \end{tabular}
\caption{Comparison on CIFAR-100. All non- exemplar-free methods use a memory buffer of 200 exemplars. The accuracies reported here are the $ACC_{TAG}$ and $ACC_{TAW}$. The methods marked with $^{\dag}$ are based on ViT.}
\label{tab:200_cifar}
\end{table*}

\begin{table*}[tb]
\centering
  \begin{tabular}{lccccc}
    \hline
    \multirow{2}{*}{\textbf{Method}}
    &\multirow{2}{*}{\textbf{Exemplar-Free}}
    &\multicolumn{2}{c}{\textbf{Tiny-ImageNet}}
    &\multicolumn{2}{c}{\textbf{ImageNet100}}\\ 
    & &Class-IL &Task-IL&Class-IL &Task-IL \\
    \hline
    Joint& \cmark& $59.38$ & $-$ & $79.18$ & $-$\\
    \hline
    ER \hfill \citep{riemer2018learning}   & \xmark
    &\phantom{0}$8.79$  &$39.16$& \phantom{0}$9.58$ & $36.24$
     \\
    AGEM \hfill \citep{chaudhry2018efficient} & \xmark 
    &\phantom{0}$8.28$  &$23.79$& \phantom{0}$9.27$ & $25.20$ 
     \\
    iCaRL \hfill \citep{rebuffi2017icarl} & \xmark
    &\phantom{0}$8.64$  &$28.41$& $12.59$ & $33.75$ 
     \\
    FDR \hfill \citep{benjamin2018measuring}  & \xmark 
    &\phantom{0}$8.77$  &$40.15$& $10.08$ & $37.80$ 
     \\
    DER++ \hfill \citep{buzzega2020dark} & \xmark 
    &$11.16$ &$40.91$& $11.92$ & $31.96$ 
     \\
    ERT \hfill \citep{buzzega2021rethinking}  & \xmark 
    &$10.85$ &$39.54$& $13.51$ & $36.94$ 
     \\
    RM \hfill \citep{bang2021rainbow}   &\xmark 
    &$13.58$ &$41.96$& $16.76$ & $35.18$ 
     \\
    PASS \hfill \citep{zhu2021prototype}
    & \cmark & 24.23&- &25.22&-\\
    SDC \hfill \citep{yu2020semantic}
    & \cmark& \phantom{0}3.94 & -&11.52 &-\\
    \hline
    LVT$^{\dag}$ \hfill \citep{wang2022continual} &\xmark 
    &$17.34$ &$46.15$& $19.46$ & $41.78$ 
     \\
    DyToX$^{\dag}$ \hfill \citep{douillard2022dytox} &\xmark 
    &$13.14$ & - & $24.82$ & - 
     \\
    A-D$^{\dag}$ \hfill \citep{pelosin2022towards}
    & \cmark & \phantom{0}6.10 & 18.02 & 10.92 & 39.13 \\
    \hline
    \textbf{GCAB}$^{\dag}$ & \cmark
    & $\mathbf{26.82}$ & $\mathbf{65.92}$& $\mathbf{40.10}$ & $\mathbf{81.82}$
    \\
    \textbf{GCAB-Fast}$^{\dag}$ & \cmark
    & ${26.44 }$ & ${65.02 }$&$36.22$ &$80.28$ 
    \\
    \hline
  \end{tabular}
\caption{Comparison on Tiny-ImageNet, and ImageNet100. All non exemplar-free methods use a memory buffer of 200 exemplars.
}
\label{tab:200_tinyimagenet}
\end{table*}

\subsection{Comparison with the State-of-the-art}
We consider two different CIL scenarios: 5 tasks and 10 tasks equally split among all classes. For CIFAR-100, tasks contain 20 classes for the 5 task scenario and 10 for the 10 task scenario. For Tiny-ImageNet and ImageNet100 we consider only the 10-task scenario, with 20 and 10 classes in each task, respectively. For CIFAR-100 and Tiny-ImageNet the images are split in $N=64$ patches. For ImageNet100 the number of patches is $N=196$. We report the task-agnostic top-1 accuracy over all the classes of the dataset after training the last task: $ACC_{TAG} = \frac{1}{N}\sum^{N}_{t=1}a_t$, where $N$ is the total number of classes in the dataset. For the task-aware scenario we report the mean accuracy over all the tasks after training the last task: $ACC_{TAW} = \frac{1}{T}\sum^{T}_{t=1}a_t$, where T is the total number of tasks.
The memory buffer for the exemplar-based methods is limited to 200 images, which is the setting proposed by~\cite{wang2022continual}.

{From Tables~\ref{tab:200_cifar}~and~\ref{tab:200_tinyimagenet} we see that our method outperforms all others in both class-incremental and task-incremental scenarios. 
Our method more than doubles the class-IL results obtained by A-D, the only other exemplar-free ViT method, on all experiments. 
Especially on the more challenging Tiny-ImageNet and the larger ImageNet100, our method obtains competitive results compared to all the other methods.
In particular, for the class incremental setting we obtain a significant improvement of 9-20\% with respect to LVT even though LVT uses 200 exemplars and we do not.  
Also notable are the improved results with respect to DyToX, which, like LVT, is based on the same architecture as ours but uses exemplars. Finally, our distilled version, which requires neither multiple forward passes, nor multiple projection networks (denoted by \emph{GCAB-Fast}),
obtains very good results only slightly below \emph{GCAB}. Note that we are using GCAB distillation with 80\% capacity usage in the distilled \emph{GCAB-Fast}.

In Figure~\ref{fig:allscenarios} we also include results for 500 exemplars. We see that our exemplar-free method obtains competitive results, even outperforming LVT and DyToX on Tiny-ImageNet and obtaining similar results as DyToX-500 on ImageNet100. Further comparison with 500 exemplars is provided in Appendix A.}

\begin{table}
    \setlength{\tabcolsep}{5pt}
    \centering
    \resizebox{1\linewidth}{!}{\begin{tabular}{ccccc}
    \hline
       \makecell{\textbf{Gated} \\ \textbf{Class} \\ \textbf{Attention}} & \makecell{\textbf{Backbone} \\ \textbf{Regularization} \\ \textbf{(PFR)}}       
       & \makecell{\textbf{Feature} \\ \textbf{Drift} \\ \textbf{Compensation}} &\makecell{\textbf{CIFAR-100}\\$\mathbf{ACC_{TAG}}$}
       &\makecell{\textbf{Tiny-ImageNet}\\$\mathbf{ACC_{TAG}}$}\\
    \hline
      & & & 11.90 & \phantom{0}7.98\\
     \cmark & & & 31.35& 22.79 \\
     \cmark & \cmark & &\phantom{0}7.96& 5.52 \\ 
     \cmark & \cmark & \cmark & 35.90& 26.82 \\
    \hline
\end{tabular}}
\caption{Ablation study on the components of our architecture for 10-task scenarios on CIFAR-100 and Tiny-ImageNet.
}
 \label{tab:ablation2}
\end{table}

\begin{table*}[!h]
\resizebox{1\linewidth}{!}{

\begin{tabular}{cccccccc}
\hline
Dataset                        & Method & Exemplar-Free   & Model Size (M params) / (MB) & Exemplars (\#) / (MB) & Total Size (MB) &\multicolumn{2}{c}{Accuracy}                     \\ \hline
\multirow{6}{*}{CIFAR100}         & GCAB & {\color{green}\checkmark}   &    12.4 / 47.30     &     0 / 0           & 47.30           &      49.86 & 35.90 \\
                               & GCAB-Fast& {\color{green}\checkmark} & 9.2 / 35.09 &0 / 0  &35.09&           48.85 &35.42                           \\
                               & Dytox &{\color{red}\xmark}    &    10.7 / 40.82    & 200 / 2.34         &  43.16                &            36.52&25.94                          \\
                               & Dytox & {\color{red}\xmark}     &    10.7 / 40.82      &  500 / 5.86    &         46.68        &                 51.28&36.24            \\
                               & LVT&{\color{red}\xmark}        &    8.9 / 33.95    &  200 / 2.34     &         36.29        &                  39.68&35.41                             \\
                               & LVT &{\color{red}\xmark}      &    8.9 / 33.95   &     500 / 5.86 &           39.81      &                  44.73&43.51                              \\ \hline
\multirow{6}{*}{Tiny-Imagenet} & GCAB&{\color{green}\checkmark}       &           12.4 / 47.30       &           0 / 0       &     47.30            & \multicolumn{2}{c}{26.82}                      \\
                               & GCAB-Fast& {\color{green}\checkmark} &            9.2 / 35.16      &      0 / 0       &       35.16  & \multicolumn{2}{c}{26.44}             \\
                               & Dytox&{\color{red}\xmark}     &            10.7 / 40.81        &  200 / 9.38       &         50.19        & \multicolumn{2}{c}{13.14}                            \\
                               & Dytox&{\color{red}\xmark}     &            10.7 / 40.81        &           500 / 23.44      &     64.25            & \multicolumn{2}{c}{24.64}                     \\
                               & LVT &{\color{red}\xmark}      &           9.0 / 34.33        &           200 / 9.38       &          43.71       & \multicolumn{2}{c}{17.34}                        \\
                               & LVT &{\color{red}\xmark}      &           9.0 / 34.33       &          500 / 23.44       &    57.77             & \multicolumn{2}{c}{23.97}                            \\ \hline
\multirow{6}{*}{ImageNet100}  & GCAB& {\color{green}\checkmark}      &           12.7 / 48.4        &           0 / 0       &       48.4          & \multicolumn{2}{c}{40.10}                            \\
                               & GCAB-Fast& {\color{green}\checkmark} &          9.5 / 36.19      &         0 / 0       &      36.19           & \multicolumn{2}{c}{36.22}            \\
                               & Dytox&{\color{red}\xmark}     &           11.0 / 41.96     &         200 / 114.84       &         156.80        & \multicolumn{2}{c}{24.82}                      \\
                               & Dytox&{\color{red}\xmark}     &           11.0 / 41.96     &         500 / 287.11       &       329.07          & \multicolumn{2}{c}{42.40}                     \\
                               & LVT&{\color{red}\xmark}       &           9.0 / 34.33      &         200 / 114.84      &     149.17            & \multicolumn{2}{c}{19.46}                       \\
                               & LVT&{\color{red}\xmark}       &           9.0 / 34.33      &         500 / 287.11      &         321.44        & \multicolumn{2}{c}{26.32}                            \\ \hline
\end{tabular}}
\vspace{-0.2cm}
\captionsetup{font={footnotesize}}
\caption{Comparison between transformer-based methods. For CIFAR-100 we report both 5- and 10-split performance. Class-incremental learning accuracy is reported.}
\label{tab:params}
\end{table*}

\subsection{Ablation Study}

We ablate on the importance of the different components of our approach. In Table~\ref{tab:ablation2} we report four possible configurations on the 10-task split of CIFAR-100 (and Tiny-ImageNet). First, we consider fine-tuning our architecture without applying any continual learning strategy to the base architecture, which results in very low performance of 11.90\% (7.98\% Tiny-ImageNet). Then we add gated class-attention to the transformer decoder. This increases the average accuracy by 20\% (15\% Tiny-ImageNet), showing the importance of preventing forgetting in the final block. 
As explained in Section~\ref{sec:gated-class-attention}, we use only two masks for gating the transformer decoder. This does not prevent backbone weight drift when passing from one task to the next. 

When we apply feature drift compensation, we obtain better performance -- notably, a more than 4\% (4\% Tiny-ImageNet) increase. These results confirm the importance of projecting the learned backbone features to the previous feature space.\footnote{We also applied feature distillation~\citep{hou2019learning} as a backbone regularization method. This does not increase performance, yielding 30.50\% (22.17\% Tiny-ImageNet).} Note that only applying PFR without feature drift compensation leads to a dramatic drop in performance to 7.96\% (5.52\% Tiny-ImageNet).

\subsection{Memory Requirements}
To further improve comparability, we report here a memory budget analysis in Table \ref{tab:params} motivated by the paper suggested by \citep{zhou2022model}.  GCAB requires just a few MB more with respect to the other methods. On Tiny-ImageNet and ImageNet100, our method is the one with the smaller computational burden while, at the same time, being the most effective. However, on CIFAR-100 with its smaller images, LVT obtains better performance at lower total memory usage (using 500 exemplars). We stress that, depending on the application, the usage of exemplars might be forbidden.

\subsection{Hyperparameter Analysis}
\begin{figure}
    \centering
    \hspace*{-0.2in}
    \includegraphics[width=\linewidth]{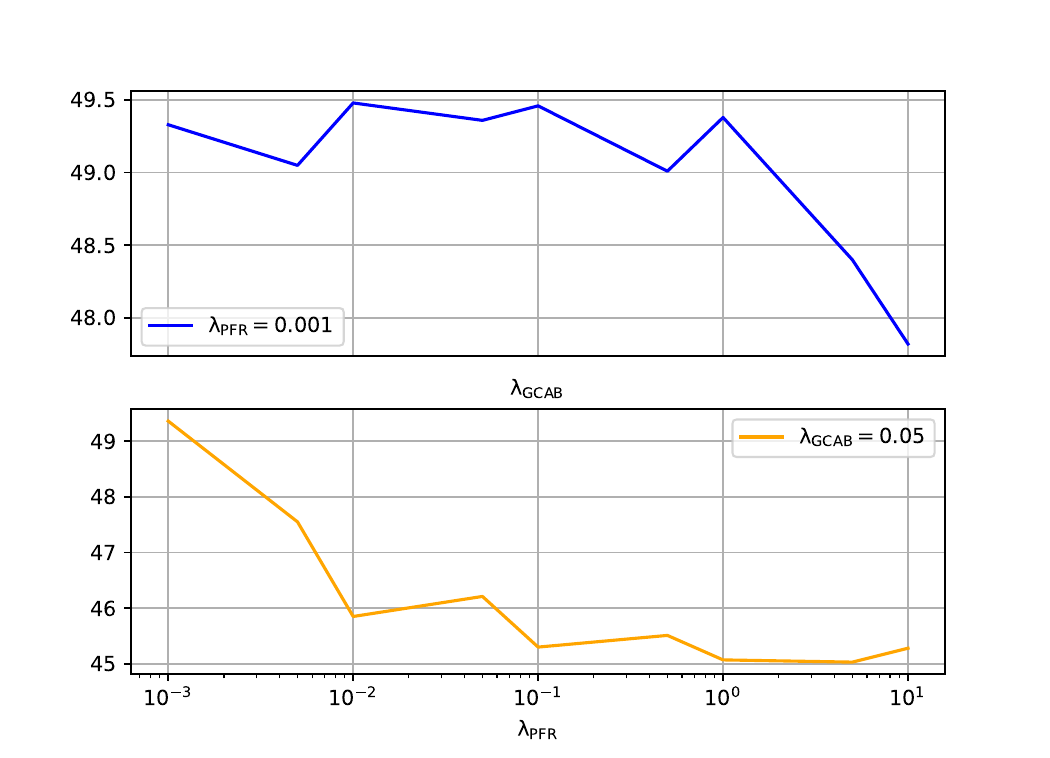}
    \vspace{-0.2cm}
    \caption{$ACC_{TAG}$ on CIFAR-100 for the 5-task scenario under varying  hyperparameters. \textbf{Top:} for fixed $\lambda_{\mathrm{pfr}}$, the $ACC_{TAG}$ as a function of $\lambda_{\mathrm{GCAB}}$. \textbf{Bottom:} for fixed $\lambda_{\mathrm{GCAB}}$, the $ACC_{TAG}$ as a function of $\lambda_{\mathrm{pfr}}$.}
    \label{fig:hyperparam}
\end{figure}

We analyze the importance and the robustness of our method with respect to the hyperparameters described is Section \ref{sec:method}. In Figure \ref{fig:hyperparam} we show the behavior of our method in terms of $ACC_{TAG}$ on the CIFAR-100 5-task scenario under changing hyperparameters. In particular, in the upper part of the figure the accuracy as a function of $\lambda_{\mathrm{GCAB}}$  is shown. In this plot, the other hyperparameter $\lambda_{pfr}$ is kept fixed  at 0.001. The accuracy is stable, confirming the robustness of our method over a wide range of values of $\lambda_{\mathrm{GCAB}}$. Only for extremely large values of $\lambda_{\mathrm{GCAB}}$ does the accuracy significantly drop.

In the lower part of Figure~\ref{fig:hyperparam} we show the variation of the $ACC_{TAG}$ as a function of $\lambda_{pfr}$ when $\lambda_{GCAB}=0.05$. For higher values of $\lambda_{pfr}$, the value of $\mathcal{L}_{\text{pfr}}$ strongly overpowers $\mathcal{L}_{\text{BCE}}$ and  $\mathcal{L}_{\text{GCAB}}$, pushing the model to not correctly learn useful features for the classification and the mask. For smaller values of $\lambda_{pfr}$ the magnitude of this term of the loss function is comparable with the others, which emphasizes the importance of correctly learning a projector network that will be used during inference.

\begin{figure*}[tb]
\begin{center}
\includegraphics[width=\linewidth]{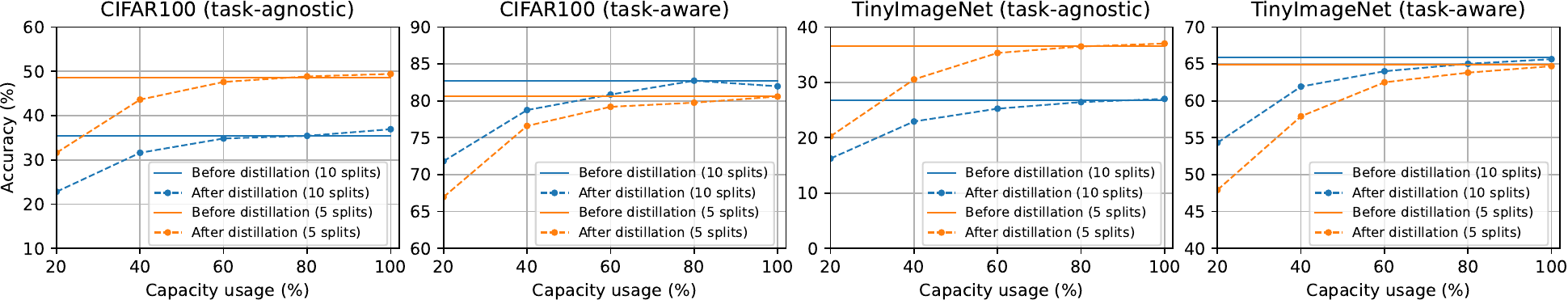}
\end{center}
\caption{Average accuracy after the last task with and without GCAB distillation.}
\label{fig:dis_res}
\end{figure*}

\subsection{Gating Capacity}

\begin{figure}[tb]
    \centering
    \includegraphics[scale=0.55]{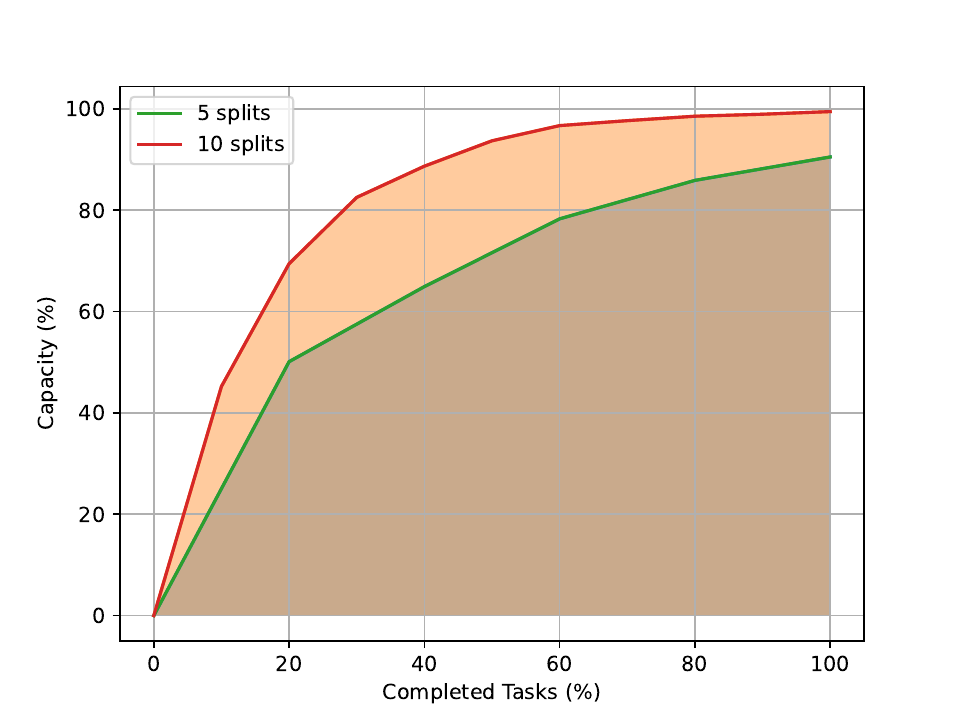}
    \caption{Gated Class-Attention Block capacity on CIFAR-100 in the 5- and 10-split scenarios. }
    \label{fig:compression}
\end{figure}
In Figure \ref{fig:compression}, we show the percentage of used masked capacity as tasks are added. The experiments were performed on CIFAR-100 dataset for 5- and 10-task scenarios. We observe that most of the available capacity is used during the firsts tasks. For subsequent ones the percentage of occupied capacity is significantly lower compared to the first ones. From the 5-task scenario curve we that there is almost 10\% of the capacity available for possible new incoming tasks. For the 10-task scenario, however, after completing 60\% of the tasks the capacity is almost full and the last tasks are using less than 2\% of capacity each.

\subsection{GCAB Distillation}
\label{sec:distillation-results}

We conduct the experiments with GCAB distillation in four scenarios: CIFAR-100 with 5 and 10 tasks, and Tiny-ImageNet with 5 and 10 tasks. We freeze the transformer encoder, projection network, GCAB, and classifiers after the last task and train the student CAB (as shown in Figure~\ref{fig:distill}) with an Adam optimizer and a learning rate $5e^{-3}$ for 200 epochs with only data from the last task.
When performing GCAB distillation, we use static binary masks at the same position as the masks in GCAB (see Figure~\ref{fig:gating_proj} (a)) to control the capacity usage in the student CAB (this potentially allows us to continue training on further tasks).

\begin{table}
    \centering
    \begin{tabular}{lccc}
    \hline
         \textbf{Method} &\textbf{Params} & \textbf{Time (ms)} & \textbf{GFLOPs} \\
    \hline
        Dytox  &10.7 & $\sim 16$ & 0.775 \\
        \textbf{GCAB} &12.4 & $\sim 22$ & 0.946 \\
        \textbf{GCAB-Fast} &9.2 & $\mathbf{\sim 7}$ &\textbf{0.510} \\
    \hline
    \end{tabular}
    \caption{Comparison of runtime and computational cost for a single forward pass (on a Quadro RTX 6000 GPU with batch size 1) for DyTox and our method with and without GCAB distillation after the last task on the CIFAR-100 10 task scenario.}
    \label{tab:dis}
\end{table}

From Figure~\ref{fig:dis_res} we see that GCAB distillation obtains almost the same performance as the model before distillation when the capacity usage is higher than 80\%, and only a small performance drop at 60\% capacity usage. As shown in Table~\ref{tab:dis}, GCAB distillation (indicated by \emph{GCAB-Fast}) can overcome the increased computational cost from multiple forward passes for each tasks during inference. 

\section{Discussion and Conclusions}
Within the context of the broader continual learning literature, we believe that our paper advances theory along two main axes. First of all, it makes a contribution to parameter-isolation methods~\citep{delange2021continual}. These methods are very popular for task-incremental learning~\citep{mallya2018packnet,mallya2018piggyback, serra2018overcoming}, but to the best of our knowledge they have not yet been applied to class-incremental learning scenarios. The main bottleneck in applying these methods to class-incremental problems is the fact that, without a task-ID available only in task-incremental scenarios, these methods require multiple forward passes through the network, which makes them computationally less attractive. In this paper, we have addressed this problem by only applying parameter isolation to the last transformer block, reducing the computational overhead to multiple forward passes only through this single transformer block. As a result, parameter isolation can be applied to class-incremental learning and only incur minor computational overhead. In addition, we show that knowledge distillation can be applied (GCAB-Fast) to replace the final block with a single block that requires only a single forward pass.

The second contribution of this paper is the cascaded feature drift compensation, which is related to semantic drift compensation~\citep{yu2020semantic}. Since the proposed gated class-attention only operates on the last block of the transformer, the backbone, which is shared among all tasks, still suffers from representation drift. To address this, rather than preventing the drift from happening, which is the prevailing approach in literature~\citep{kirkpatrick2017overcoming, li2017learning}, we accept that drift will occur when learning the new task and compensate for it. We show that learned projection networks (which are used to ensure that the new network consolidates the knowledge of the previous one) allow for a straightforward compensation of representation drift. Differently than the method proposed in~\citep{yu2020semantic}, our approach is not limited to prototype-based (nearest class mean) classification and can also be applied to multi-layer classifiers, like the applied class-attention block used in our paper. 

In conclusion, we presented an exemplar-free approach to class-incremental visual transformer training. Our method, through the gated class-attention mechanism,  achieves low forgetting by learning and masking important neurons for each task. High plasticity is ensured via backbone regularization and feature drift compensation using a cascade of feature projection networks. Experiments on several benchmark datasets show that our method obtains competitive results when compared to rehearsal based ViT methods. Ours is one of the first effective approaches to exemplar-free, class-incremental training of ViTs. 

\section*{Acknowledgments}
We acknowledge the project, TED2021-132513B-I00, PID2022-143257NB-I00 (MICINN, Spain), the CERCA Programme of Generalitat de Catalunya and the European Commission funded Horizon 2020 project, grant number 951911 (AI4Media). 

\section*{Data Availability}
In the interest of reproducibility, we have made our code available at \url{https://github.com/OcraM17/GCAB-CFDC}. All experiments are conducted on publicly available datasets; see the references cited.

\bibliographystyle{spbasic}      %
\bibliography{manuscript}

\appendix

\section{Comparison with exemplar-based methods using 500 exemplars}
As a further comparison we analyzed the behavior of GCAB compared to exemplar-based methods with a replay buffer of 500 exemplars. In Table~\ref{tab:500_cifar} we report the results of the considered methods on the CIFAR-100 dataset in 5- and 10-task scenarios. In these settings the 100 classes have been split in 5 and 10 task, each one containing  20 and 10 classes, respectively. The performance of both our \textit{exemplar-free} solutions (GCAB and GCAB-Fast) are comparable with exemplar-based methods using 500 exemplars. In detail, on the 5-task scenario, our method reaches the second-best result. Similarly, on the 10-task scenario, GCAB and GCAB-Fast achieve very good performance at the third and fourth positions in the comparison.

\begin{table*}[tb]
\centering
  \begin{tabular}{lccccc}
    \hline
    \multirow{2}{*}{\textbf{Method}} 
    &\multirow{2}{*}{\textbf{Exemplar-Free}}
    &\multicolumn{2}{c}{\textbf{CIFAR-100 5 Tasks}}
    &\multicolumn{2}{c}{\textbf{CIFAR-100 10 Tasks}}\\
    & &Class-IL &Task-IL &Class-IL &Task-IL\\
    \hline
    Joint& \cmark& $75.39$ & - & $75.39$ & -\\
    \hline
    ER \hfill \citep{riemer2018learning}   & \xmark&$27.97$& $68.21$& $21.54$& $74.97$\\
    GEM \hfill \citep{lopez2017gradient} & \xmark &$25.44$& $67.49$& $18.48$& $72.68$
    \\
    AGEM \hfill \citep{chaudhry2018efficient} & \xmark &$18.75$ &$58.70$ &$9.72$  &$58.23$ 
     \\
    iCaRL \hfill \citep{rebuffi2017icarl} &\xmark&$35.95$ &$64.40$ &$30.25$ &$71.02$ 
     \\
    FDR \hfill \citep{benjamin2018measuring}  & \xmark &$29.99$ &$69.11$ &$22.81$ &$74.22$ 
     \\
    GSS \hfill \citep{aljundi2019gradient}  & \xmark &$22.08$ &$61.77$ &$13.72$ &$56.32$ 
     \\
    DER++ \hfill \citep{buzzega2020dark} & \xmark &$38.39$ &$70.74±$ &$36.15$ &$73.31$ 
     \\
    HAL \hfill \citep{chaudhry2021using}  & \xmark &$16.74$ &$39.70$ &$11.12$ &$41.75$ 
    \\
    ERT \hfill \citep{buzzega2021rethinking}  & \xmark &$28.82$ &$62.85$ &$23.00$ &$68.26$ 
     \\
    RM \hfill \citep{bang2021rainbow}   &\xmark &$39.47$ &$69.27$ &$32.52$ &$73.51$ 
     \\
    PASS \hfill \citep{zhu2021prototype}
     & \cmark & $48.28$ & - & $33.76$ & -\\
    SDC \hfill \citep{yu2020semantic}
    &  \cmark& 6.65 &-&7.41&-\\
    \hline
    LVT$^{\dag}$ \hfill \citep{wang2022continual} &\xmark &$44.73±$ &$71.54$ &$\mathbf{43.51}$ &$76.78$  
     \\
    DyToX$^{\dag}$ \hfill \citep{douillard2022dytox} &\xmark & $\mathbf{51.28}$ & -  & 36.24 & -  
     \\
    A-D$^{\dag}$ \hfill \citep{pelosin2022towards}
    &  \cmark & 15.61 & 42.82 & $16.77$& $55.53$\\
    \hline
    \textbf{GCAB}$^{\dag}$ & \cmark& $49.86$ & $\mathbf{81.01 }$ & $35.90 $ & $\mathbf{82.08}$ 
    \\
    \textbf{GCAB-Fast}$^{\dag}$ & \cmark & $48.85 $ & $79.75 $ & $35.42$ & $81.97$ 
    \\
    \hline
  \end{tabular}
\caption{Comparison on CIFAR-100. All exemplar-based methods use a memory buffer of 500 exemplars. The accuracies reported here are the $ACC_{TAG}$ and $ACC_{TAW}$ computed after training the last task. Methods marked with $^{\dag}$ are based on ViT.}
\label{tab:500_cifar}
\end{table*}

In Table~\ref{tab:500_tiny}, we report a comparison on Tiny-Imagenet and ImageNet100. For these two datasets, we consider the 10 and 20-task scenario, where each task contains 20 and 10 classes, respectively. Like in Table~\ref{tab:500_cifar}, the exemplar-based methods use a buffer of 500 exemplars. The performance of GCAB and GCAB-Fast on the Tiny-ImageNet surpass all exemplar-based methods even with a budget of 500 exemplars. In ImageNet100, our method is competitive with exemplar-based methods, reaching the second-best accuracy after DyToX with 500 exemplars.

\begin{table*}[tb]
\centering
  \begin{tabular}{lccccc}
    \hline
    \multirow{2}{*}{\textbf{Method}} 
    &\multirow{2}{*}{\textbf{Exemplar-Free}}
    &\multicolumn{2}{c}{\textbf{Tiny-ImageNet}}
    &\multicolumn{2}{c}{\textbf{ImageNet100}}\\ 
    & &Class-IL &Task-IL&Class-IL &Task-IL \\
    \hline
    Joint& \cmark& $59.38$ & $-$ & $79.18$ & $-$\\
    \hline
    ER \hfill \citep{riemer2018learning}  & \xmark
    &$10.15$  &$50.11$& $11.68$ & $42.04$
     \\
    AGEM \hfill \citep{chaudhry2018efficient} & \xmark 
    &$9.67$  &$26.79$&$10.92$ & $34.22$ 
     \\
    iCaRL \hfill \citep{rebuffi2017icarl} & \xmark
    &$10.69$  &$35.89$& $16.44$ & $36.89$ 
     \\
    FDR \hfill \citep{benjamin2018measuring}  & \xmark 
    &$10.58$  &$49.91$& $11.78$ & $42.60$ 
     \\
    DER++ \hfill \citep{buzzega2020dark} & \xmark 
    &$19.33$ &$51.90$& $14.52$ & $35.46$ 
     \\
    ERT \hfill \citep{buzzega2021rethinking}  & \xmark 
    &$12.13$ &$50.87$& $20.42$ & $41.56$ 
     \\
    RM \hfill \citep{bang2021rainbow}   &\xmark 
    &$18.96$ &$52.08$& $14.56$ & $38.66$ 
     \\
    PASS \hfill \citep{zhu2021prototype}
     & \cmark & 24.23&- &25.22&-\\
    SDC \hfill \citep{yu2020semantic}
     & \cmark&3.94 & -&11.52 &-\\
    \hline
    LVT$^{\dag}$ \hfill \citep{wang2022continual} &\xmark 
    &$23.97$ &$57.39$& $26.32$ & $47.84$ 
     \\
    DyToX$^{\dag}$ \hfill \citep{douillard2022dytox} &\xmark 
    &24.64 & - & $\mathbf{42.40}$ & - 
     \\
    A-D$^{\dag}$ \hfill \citep{pelosin2022towards}
     & \cmark & 6.10 & 18.02 & 10.92 & 39.13 \\
    \hline
    \textbf{GCAB}$^{\dag}$ & \cmark
    & $\mathbf{26.82 }$ & $\mathbf{65.92 }$& $40.10$ & $\mathbf{81.82}$
    \\
    \textbf{GCAB-Fast}$^{\dag}$& \cmark
    & ${26.44 }$ & ${65.02 }$&$36.22$ &$80.28$ 
    \\
    \hline
  \end{tabular}
\caption{Comparison on Tiny-ImageNet and ImageNet100. All exemplar-based methods use a memory buffer of 500 exemplars. The accuracies reported here are the $ACC_{TAG}$ and $ACC_{TAW}$ computed after the training the last task. Methods marked with $^{\dag}$ are based on ViT.} 
\label{tab:500_tiny}
\end{table*}

\section{Additional Embedding Visualizations}
\begin{figure*}[h]
\begin{center}
\includegraphics[scale=0.4]{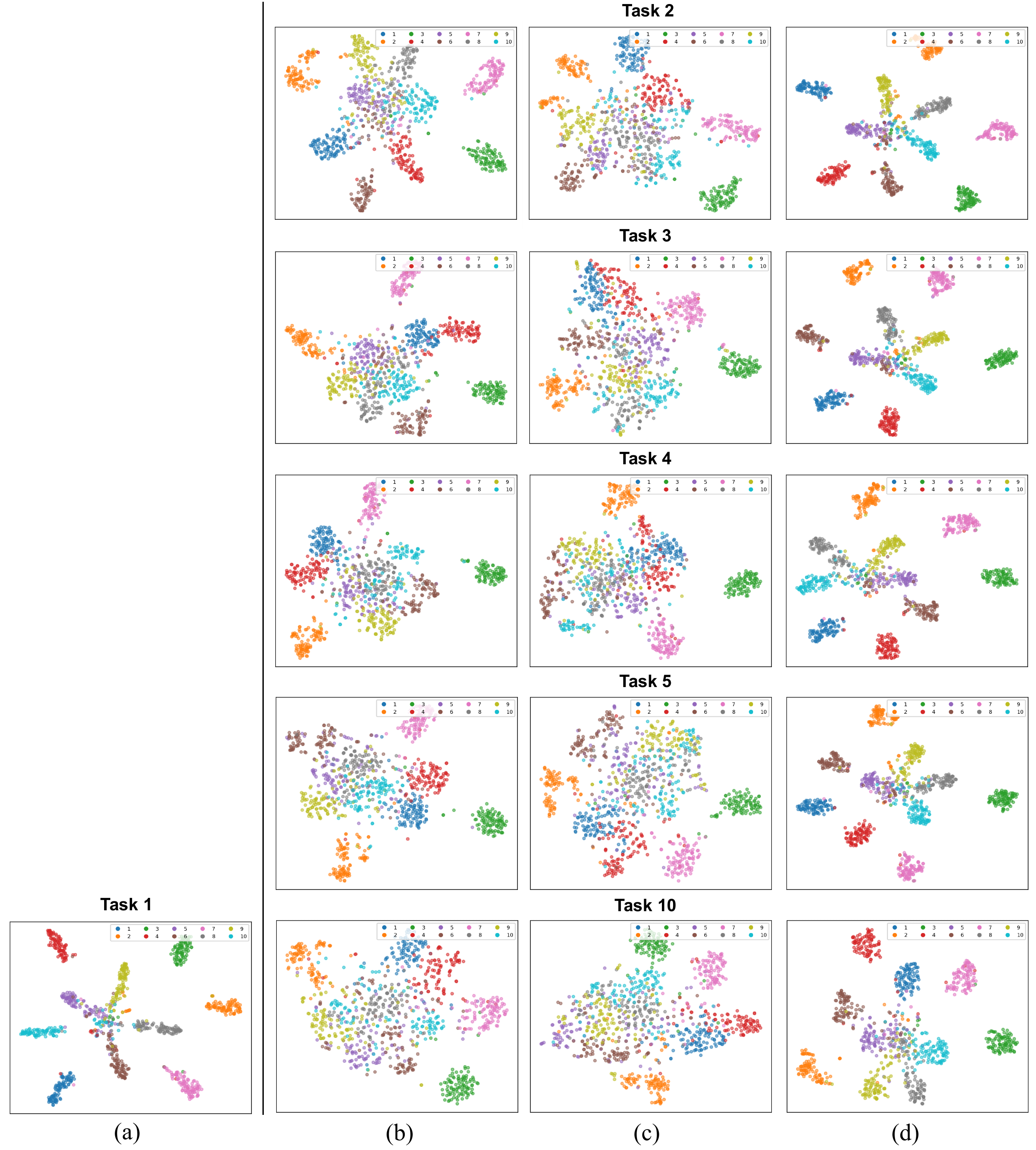}
\end{center}
\caption{t-SNE visualization of embedding space (output of $f(\cdot)$) at Tasks 1, 2, 3, 4, 5, and 10. (a) Results after Task 1; (b) Results without PFR during training; (c) Results with PFR during training but without feature drift compensation; (d) Results with PFR during training and with feature drift compensation.}
\label{fig:embedding_tag}
\end{figure*}

We give here visualizations of the embeddings produced with and without the projection network cascade. In column (a) of Figure~\ref{fig:embedding_tag}, we show the embedding obtained after training the first task. In column (b) we report the result of the embedding without PFR during training. In columns (c) and (d) we show the embeddings obtained using  PFR during training. The difference between (c) and (d) is the use of the Feature Drift Compensation mechanism during inference. The projection networks of the various tasks are used to align the embeddings produced by the current backbone to the ones produced by the previous ones (in column d). The different rows of the figure show the embeddings of the task 1 data produced by the network after training a variable number of tasks. 
We see that the embeddings of Task 1 without PFR during training (b) and only with PFR during training but not during inference (c) are less clearly clustered, especially after Task 10 where samples of all classes significantly overlap. However, when using PFR during both training and inference the data remains clearly clustered, as seen in column (d).

\end{document}